\documentclass[journal]{IEEEtran}
\usepackage{cite}

\usepackage{graphicx}
\ifCLASSINFOpdf
\else
\fi

\usepackage[colorlinks,
           linkcolor=red,
           anchorcolor=blue,
           citecolor=green
           ]{hyperref}
\begin{document}
%
\title{Multimodal Intelligence: Representation Learning, Information Fusion, and Applications}
%
%
%

\author{Chao Zhang,~
        Zichao Yang,~
        Xiaodong He,~\IEEEmembership{Fellow,~IEEE,}
        and~Li~Deng,~\IEEEmembership{Fellow,~IEEE}
\thanks{C. Zhang and X. He were with JD AI Research. C. Zhang was also with the Department of Engineering, University of Cambridge, UK.}
\thanks{Z. Yang and L. Deng were with Citadel LLC.}
\thanks{Corresponding author: Xiaodong He $<$\texttt{xiaodong.he@jd.com}$>$}
}

%
%

\markboth{To Appear in IEEE Journal of Selected Topics in Signal Processing}%
{Shell \MakeLowercase{\textit{et al.}}: Bare Demo of IEEEtran.cls for IEEE Journals}
%



\maketitle

\begin{abstract}
Deep learning methods have revolutionized speech recognition, image recognition, and natural language processing since 2010. Each of these tasks involves a single modality in their input signals. However, many applications in the artificial intelligence field involve multiple modalities. Therefore, it is of broad interest to study the more difficult and complex problem of modeling and learning across multiple modalities. In this paper, we provide a technical review of available models and learning methods for multimodal intelligence. The main focus of this review is the combination of vision and natural language modalities, which has become an important topic in both the computer vision and natural language processing research communities. 

This review provides a comprehensive analysis of recent works on multimodal deep learning from three perspectives: learning multimodal representations, fusing multimodal signals at various levels, and multimodal applications. Regarding multimodal representation learning, we review the key concepts of embedding, which unify multimodal signals into a single vector space and thereby enable cross-modality signal processing. We also review the properties of many types of embeddings that are constructed and learned for general downstream tasks. Regarding multimodal fusion, this review focuses on special architectures for the integration of representations of unimodal signals for a particular task. Regarding applications, selected areas of a broad interest in the current literature are covered, including image-to-text caption generation, text-to-image generation, and visual question answering. We believe that this review will facilitate future studies in the emerging field of multimodal intelligence for related communities.
\end{abstract}


%
\IEEEpeerreviewmaketitle

\section{Introduction}
\label{sec:intro}
%
%
%
%
\IEEEPARstart{S}{ignificant} progress has been made in the field of machine learning in recent years based on the rapid development of deep learning algorithms \cite{Hinton2006:Science,Bengio2009:FnT,Deng2014:FnT,Schmidhuber2015:NN,LeCun2015:Nature,Goodfellow2016:Book}. 
The first major milestone was a significant increase in the accuracy of large-scale automatic speech recognition based on the use of fully connected deep neural networks (DNNs) and deep auto-encoders around 2010 \cite{YuDeng2010,DengHinton2010,Deng2011,YuPatent2011,Dahl2011,Deng2013:Microsoft,Dahl2015:TASLP,Seide2011:Interspeech,Hinton2012:SigProc,DengHinton2013,YuDeng2015}. Shortly thereafter, a series of breakthroughs was achieved in computer vision (CV) using deep convolutional neural network (CNN) models \cite{LeCun1998:ProcIEEE} for large-scale image classification around 2012 \cite{Krizhevsky2012:AlexNet,Simonyan2015:VGGNet,He2016:ResNet,Szegedy2015:Inception} and large-scale object detection around 2014 \cite{Girshick2014:RCNN,Girshick2015:FastRCNN,Ren2015:FasterRCNN}. All of these milestones have been achieved for pattern recognition with a single input modality.
In natural language processing (NLP), recurrent neural network (RNN) based semantic slot filling methods \cite{Mesnil2015:SlotFilling} have achieved state-of-the-art for spoken language understanding.  RNN-encoder-decoder models with attention mechanisms \cite{Bahdanau2015:NMT}, which are also referred to as sequence-to-sequence models \cite{Sutskever2014:Seq2Seq}, have achieved superior performance for machine translation in an end-to-end fashion \cite{Wu2016:MT,Luong2015:MT}. 
For additional NLP tasks with small amounts of training data, such as question answering (QA) and machine reading comprehension, generative pre-training has achieved state-of-the-art results \cite{Peters2018:Elmo,Radford2018:GPT,Devlin2019:BERT}. This method transfers parameters from a language model (LM) pre-trained on a large out-of-domain dataset using unsupervised training or self-training, which is followed by fine-tuning on small in-domain datasets.

Although there have been significant advances in vision, speech, and language processing, many problems in the artificial intelligence field involve more than one input modality, such as intelligent personal assistant systems that must understand human communication based on spoken words, body language, and pictorial languages \cite{Shum2018:FITEE}. Therefore, it is of broad interest to study modeling and training approaches across multiple modalities \cite{BengioDeng}. Based on advances in image processing and language understanding \cite{DengLiu}, tasks combining images and text have attracted significant attention, 
including visual-based referred expression understanding and phrase localization \cite{Kazemzadeh2014:match,Yu2016:match,Plummer2015:match}, as well as image and video captioning \cite{karpathy2015deep,Vinyals2015:cvpr,Johnson2016:cvpr,Xu2016:cvpr,Pan2016:cvpr,You2016:cvpr}, visual QA (VQA) \cite{item86,item87,item94}, text-to-image generation \cite{Yan2016:eccv,Reed2016:icml,AttnGAN}, and visual-and-language navigation \cite{Anderson2018:cvpr}. In these tasks, natural language plays a key role in helping machines in ``understanding'' the content of images, where ``understanding'' means capturing the underlying correlations between the semantics embedded in languages and the visual features obtained from images. In addition to text, vision can also be combined with speech to perform audio-visual speech recognition \cite{Dupont2000:TMM,Cookea2006:JASA,Afouras2018:TPAMI}, speaker recognition \cite{Maison1999:MMSP,Wu2005:Speaker,Chung2018:Interspeech}, speaker diarization, \cite{Gebru2018:TPAMI,Chung2019:Interspeech}, as well as speech separation \cite{Wu2019:ASRU,Ephrat2018:TOG} and enhancement \cite{Afouras2018:Interspeech}.

This paper provides a technical review of the models and training methods used for multimodal intelligence. Our main focus is the combination of CV and NLP, which has become an important area for both of these research communities that covers many different tasks and technologies. To provide a structured perspective, we have organized this technical review according to three key topics: representations, fusion, and applications. 
\begin{itemize}
\item Learning representations for input data is a core problem in deep learning. For multimodal tasks, collecting parallel data across different modalities can be a difficult task. Leveraging pre-trained representations with the desired properties, such as properties suitable for zero-shot or few-shot learning, is often an effective solution to this issue. Both supervised and unsupervised training-based multimodal representation learning methods are reviewed.  
\item Fusing the representations of different modalities is a core problem in any multimodal task. Unlike previous studies that have classified related work based on the stage in which fusion occurs within a procedure, we classify related work according to the actual operations used during fusion, such as attention mechanisms and bilinear pooling, because it is difficult to classify recent complex approaches based on stages.     
\item Three types of applications are reviewed: image captioning, text-to-image generation, and VQA. These applications provide examples of how representation learning and fusion can be applied to specific tasks, as well as a representation of the current development of multimodal applications, particularly those integrating vision with natural languages. Visual reasoning methods are also discussed.
\end{itemize}

The remainder of this paper is organized as follows. Section~\ref{sec:representations} reviews recent progress in terms of developing representations for single or multiple modalities. Section~\ref{sec:fusion} introduces commonly used fusion methods with a focus on attention mechanisms and bilinear pooling. Section~\ref{sec:applications} introduces applications, including caption generation, text-to-image generation,  VQA, and visual reasoning, followed by a summary and our outlook regarding potential future research directions.

\section{Representations}
\label{sec:representations}
Deep learning, as a special area within representational learning, focuses on the use of artificial neural networks (ANN) with many hidden layers to discover suitable representations or features from raw data automatically for specific tasks \cite{Bengio2013:TPAMI}. Representation learning has great value in practice since better representations can often simplify subsequent learning tasks. Over the past decade, it has become feasible to learn effective and robust representations for single modalities, such as text \cite{Huang2013:DSSM,Shen2014:CDSSM,Palangi2016:TASLP,Peters2018:Elmo,Radford2018:GPT,Devlin2019:BERT,Rumelhart1986:Nature,Bengio2003:JMLR,Mikolov2013:ICLR,Mikolov2013:NIPS,Bojanowski2017:Nature} or images \cite{Krizhevsky2012:AlexNet,Simonyan2015:VGGNet,He2016:ResNet,Szegedy2015:Inception,Girshick2014:RCNN,Girshick2015:FastRCNN,Ren2015:FasterRCNN}, based on the availability of large amounts of data and the development of deep learning. Although multimodal representations are attracting increasing attention, they still remain a challenging problem due to the complex cross-modal interactions and possible mismatches between training and test data in each modality. 

In this section, commonly used types of single-modal representations, such as text and images, are reviewed. These representations often serve as cornerstones for learning multimodal representations.
Next, both supervised and unsupervised methods for learning a joint representation space for multiple modalities are introduced. To enable models to handle data samples with missing modalities, the zero-shot learning problem can be solved to increase the similarity of representational spaces across the involved modalities. 
Finally, inspired by the success of adapting pre-trained LMs to downstream tasks in NLP, methods that leverage large unimodal datasets to improve the learning of multimodal representations are also discussed.

\subsection{Unimodal Embeddings}
A distributed representation is a vector that distributes information related to a concept with multiple elements, indicating that elements can be tuned separately to allow more concepts to be encoded efficiently in a relatively low-dimensional space \cite{Rumelhart1986:Nature}. 
Such representations can be compared to symbolic representations, such as one-hot encoding, which uses an element with a value of one to indicate the presence of a concept locally and values of zero for other elements. In deep learning, the term ``embedding'' often refers to a mapping from a one-hot vector representing a word or image category to a distributed representation of real-valued numbers. 

\subsubsection{Visual representations}
Image embeddings can be acquired as output values from the final CNN layers in models that classify images into categories, such as AlexNet \cite{Krizhevsky2012:AlexNet}, VGGNet \cite{Simonyan2015:VGGNet}, GoogLeNet \cite{Szegedy2015:Inception}, and ResNet \cite{He2016:ResNet}. 
AlexNet, GoogLeNet, and ResNet were the winners of the 2012, 2014, and 2015 ImageNet Large Scale Visual Recognition Competition for image classification, respectively \cite{deng2009imagenet,Russakovsky2015:ImageNet}. Alternatively, features with more direct relationships to semantics can be used as visual embeddings, such as convolutional features and associated class labels from selected regions identified by object detection models. Models using this approach include the region-based CNN (R-CNN) \cite{Girshick2014:RCNN}, Fast R-CNN \cite{Girshick2015:FastRCNN}, and Faster R-CNN \cite{Ren2015:FasterRCNN}. It should be noted that these models are only a few examples and do not cover all popular CNN structures. 

\subsubsection{Language representations}
Text embeddings can be derived from a neural network language model (NNLM) \cite{Bengio2003:JMLR}, which estimates the probability of a text sequence by factorizing the sequence into word probabilities using a chain rule for probability.
RNN-based NNLMs, such as long short-term memory (LSTM) or gated recurrent unit (GRU) LMs \cite{Hochreiter1997:LSTM,Chung2014:GRU}, facilitate the use of information from all past words stored in a fixed-length recurrent vector when predicting a current word. 
In addition to NNLMs, the continuous bag-of-words model, skip-grams, 
and global vectors (GloVe) are other commonly used methods for word embeddings \cite{Mikolov2013:ICLR,Pennington2014:Glove}.
A series of deep structured semantic models (DSSMs) have been proposed since 2013 for sentence-level embedding learning based on the optimization of semantic similarity-driven objectives using various neural network structures in pseudo-Siamese network settings \cite{Huang2013:DSSM,Shen2014:CDSSM,Palangi2016:TASLP,Elkahky2015:MVDSSM,Liu2015:MTDSSM,Yih2014:KBQADSSM,Yih2015:KBQADSSM}.

Recently, to accomplish downstream natural language understanding tasks with small amounts of training data, many studies have focused on learning general text embeddings by predicting word probabilities using NNLMs with complex structures based on large text corpora. Embeddings from language models \cite{Peters2018:Elmo} use combined embeddings from multiple layers of bidirectional LSTMs for forward and backward propagation. Generative pre-training \cite{Radford2018:GPT} and bidirectional encoder representations for transformers (BERT) \cite{Devlin2019:BERT} use the decoder and encoder components of transformer models to estimate the probability of a current subword unit. 
Besides the word and subword levels, text embeddings can also be learned on the phrase, sentence, and paragraph levels \cite{Kiros2015:NIPS,Sutskever2014:Seq2Seq}. 

\subsubsection{Vector arithmetic for word and image embeddings}
\label{ssec:representation:vector}
It is well known word embeddings can capture both syntactic and semantic regularities. A famous example showed that
the operation vector(``King'')$-$vector(``Man'')$+$vector(``Woman'')
results in a vector closest to the vector(``Queen''), where vector($\cdot$) denotes the representation of a word learned by an RNN LM \cite{Mikolov2013:NAACLHLT}. A similar phenomenon has also been observed for vision embeddings. 
When using a generative adversarial network (GAN)
\cite{Goodfellow2014:NIPS}, it has been shown that the operation vector(``Man with glasses'')$-$vector(``Man'')$+$vector(``Woman'') results in
vector(``Woman with glasses'') \cite{Radford2016:ICLR}, where vector($\cdot$) refers to the representation of an image. This result indicates that GANs can capture image representations that disentangle the concept of gender from that of wearing glasses. These findings regarding both text and image representations have encouraged additional studies on joint representations of these two modalities. Additional details regarding GAN-based image generation can be found in Section~\ref{ssec:application:synthesis}.

\subsubsection{Speaker representations}
\label{ssec:representation:speaker}

Despite speech-related studies are not the focus of this paper, a brief discussion on speaker representation is presented here since it is broadly used in many downstream tasks nowadays \cite{BUT2011:ASRU,Saon2013:ASRU,Senior2014:ICASSP,Yella2015:Interspeech,Wang2018:ICASSP,Wu2015:Interspeech,Wan2017:ASRU,WaveNet,GBTTS,Xia2016:Interspeech,Dan2018:Interspeech,Pappagari:2020}. 
The $i$-vector approach estimates a vector for every speaker using factor analysis \cite{iVector}. Speaker-specific vectors can be jointly trained with the DNN acoustic models \cite{JiangHui,Strom}. Speaker embeddings can also be derived as the outputs from the penultimate layer of a DNN trained to classify the training set speakers at frame level, namely the $d$-vectors \cite{dVector}.  Alternatively, the model can be trained to discriminate speakers at utterance level using a statistical pooling layer or a  self-attentive layer \cite{BoWen,Zhu2018:Interspeech,cVector}, and $x$-vector is the first of such approaches. 
The training set for $x$-vectors is often augmented to include different background noises and channels \textit{etc.}, which helps to disentangle the concept of the speaker's voice characteristics from the others.  Privacy-preserving is an important issue for speech product in practice,  secure binary embeddings can be used to estimate speaker embeddings without exposing the speaker data \cite{Isabel1,Isabel2}.

\subsection{Multimodal Representations}

Although significant progress has been made in terms of learning representations for vision or languages, it is theoretically insufficient to model a complete set of human concepts using only unimodal data. For example, the concept of a ``beautiful picture'' is grounded in visual representation, so it can be difficult to describe this concept using natural language or other non-visual approaches. Therefore, it is important to learn joint embeddings to leverage the complementarity of multimodal data to represent such concepts more accurately. 


\subsubsection{Unsupervised training methods}
Joint embeddings for multimodal data can be learned by simply reconstructing raw inputs using multiple streams of deep Boltzmann machines or auto-encoders with shared layers acting as a shared representation space \cite{item1,item2,item3}. 
Alternatively, based on the development of methods for single modalities, a shared representation space can be constructed by mapping pre-trained representation spaces for the involved individual modalities into a common space. For example, Fang \textit{et al}. proposed a deep multimodal similarity model (DMSM) \cite{fang2015captions} as an extension of the text modal DSSM to learn embedding representations of text and images in a unified vector space. The simple fusion of word and image embeddings was accomplished using addition or concatenation \cite{item3,item4}. The similarity between textual and visual embeddings can be increased through training \cite{item5}. In a recent study, the correlation and mutual information between embeddings of different modalities were maximized \cite{item6,item7}. Similarly, the distances between word embeddings can be modified according to the similarities between their visual instantiations \cite{item8}, which are determined by clustering abstract scenes in an unsupervised manner.  

Other studies have correlated image regions/fragments with sentence fragments or attribute words to generate fine-grained multimodal embeddings \cite{item9} by calculating the alignments between images and sentence fragments automatically. Wu \textit{et al.} unified the embeddings of concepts at different levels, including objects, attributes, relationships, and full scenes \cite{item10}. 
The stacked cross attention network was proposed to learn fine-grained word and image-object aligned embeddings for image-text matching \cite{Lee2018:ECCV}.
Additionally, the deep attentional multimodal similarity model (DAMSM) was proposed \cite{AttnGAN} as an extension of DMSM with attention models to measure the similarity between image sub-regions and words as an additional loss function for text-to-image generation.


\subsubsection{Supervised training methods}
Supervised training can be used to improve the learning of multimodal representations. Representations can be factorized into two sets of independent factors: multimodal discriminative factors for supervised training and intra-modality generative factors for unsupervised training \cite{item11}. Discriminative factors are shared across all modalities and are useful for discriminative tasks, whereas generative factors can be used to reconstruct missing modalities. Based on detailed text annotations, some researchers have proposed learning word embeddings from their visual co-occurrences (ViCo) when considering natural scene images or image regions \cite{item12}. The concept of ViCo is complementary to GloVe text embedding in that it more accurately represents similarities and differences between visual concepts that are difficult to obtain from text corpora alone. Multiple supervised training tasks have been applied to different layers of vision-language encoders \cite{item13}. The order of training tasks is determined based on the concept of curriculum learning to increase the complexity of training objectives in a step-by-step manner. 

\subsubsection{Methods for zero-shot learning}
Zero-shot learning is often applied to vision-related tasks based on the difficulty of acquiring sufficient labelled images for training for all possible object categories. However, not all types of multimodal representations are suitable for zero-shot learning because certain representations may require pairwise data from different modalities simultaneously. Here, we review methods that rely on additional language sources to overcome this issue. 
Deep learning-based zero-shot learning begins by developing a linear mapping layer between different pre-trained embeddings \cite{item14,item15}. The deep visual-semantic embedding model was constructed using skip-gram text embeddings and AlexNet visual features. This model allows both types of pre-trained models to be jointly trained through a linear mapping layer \cite{item15}. This model was subjected to large-scale test with 1000 known classes and 2000 unknown classes. Better representations could be learned for one-shot and few-shot image retrieval
when correlated auto-encoders were used to reconstruct the representations for each modality \cite{item16}.
A recent work used word labels related and unrelated to a target class to derive visual embeddings from a pre-trained VGG network as positive and negative visual priors, which were used as the inputs for another model to achieve the semantic image segmentation of new object classes that were unseen in the training set \cite{Golub2020:WACV}.
Rich information sources can be used for multiple modalities, including words selected from Wikipedia articles and features derived from multiple CNN layers \cite{item17}. Rather than using direct text attribute inputs, sentence embeddings generated by recurrent models can be used as a text interface for zero-shot learning to achieve enhanced results \cite{item18}. 

\subsubsection{Transformer-based methods}
\label{ssec:representation:transformer}
Transformers are prevalent sequence-based encoder-decoder models that are formed by stacking many blocks of feedforward layers with multi-head self-attention models, whose parameters are shared temporally \cite{Vaswani2017:NIPS}. 
Compared to RNN-based encoder-decoder models \cite{Bahdanau2015:NMT}, 
such models can provide superior performance on long sequences based on the removal of the first-order Markovian assumption imposed on RNNs. 
BERT, which is the encoder component of a transformer pre-trained on a large text corpus as a masked LM, is a standard choice for text embeddings for downstream tasks. 
Therefore, it is natural to generalize text-only BERT to cover images and derive pre-trained bimodal embeddings.  


A straightforward method for extending unimodal BERT to bimodal applications is to include new tokens to indicate visual feature inputs, such as those proposed in \cite{item23,item24,item25,item26,item58}. Additionally, the transformer model can be modified by introducing an extra encoder or attention structures for visual features \cite{item21,item22,item20}. Additional details regarding modified structures can be found in Section~\ref{ssec:fusion:attention}. Furthermore, a recent NLP study suggested that multitask learning can improve the generalization ability of BERT representations \cite{Liu2019:ACL}. 
Therefore, most of the aforementioned bimodal BERT-based models adopt multitask training to improve their performance on downstream tasks, such as VQA, and image and video captioning \textit{etc}. 

\section{Fusion}
\label{sec:fusion}
Fusion is a key research topic in multimodal studies, which integrates information extracted from different unimodal data sources into a single compact multimodal representation. There is a clear connection between fusion and multimodal representations. We classify an approach into the fusion category if it focuses on architectures for integrating unimodal representations for a particular task.
Fusion methods can be divided based on the stage in which fusion occurs during the associated procedures. Because early and late fusion can suppress either intra- or inter-modality interactions, recent studies have focused on intermediate methods that allow fusion to occur on multiple layers of a deep model. 

This section presents a review of intermediate fusion not only because it is more flexible but also because the boundaries between stages are less clear based on the use of unimodal features derived from pre-trained backbone models. Three types of methods that are mainly used to fuse text with image features are considered, namely, simple operation-based, attention-based, and tensor-based methods. 

\subsection{Simple Operation-based Fusion}
In deep learning, vectorized features from different information sources can be integrated using simple operations, such as concatenation or weighted sums, which often have few or no associated parameters because the joint training of deep models can adapt layers for high-level feature extraction to adjust to the required operations. 
\begin{itemize}
\item Concatenation can be used to combine either low-level input features \cite{item31,item32,item33} or high-level features extracted by pre-trained models \cite{item33,item34,item35}.
\item For weighted sums with scalar weights, an iterative method that requires the pre-trained vector representations to have the same number of elements arranged in an order that is suitable for element-wise addition has been proposed \cite{item36}. This can be achieved by training a fully connected layer for dimension control and reordering for each modality.
\end{itemize}
 A recent study \cite{item37} employed neural architecture search with progressive exploration \cite{item38,item39,item40} to find suitable settings for a number of fusion functions. Each fusion function was configured in terms of which layers to fuse and whether to use concatenation or a weighted sum as the fusion operation. 

\subsection{Attention-based Fusion}
\label{ssec:fusion:attention}
Attention mechanisms are widely used for fusion. Attention mechanisms often refer to the weighted sum of a set of vectors with scalar weights that are dynamically generated by a small ``attention'' model at each time-step \cite{item41,item42}. Multiple glimpses (output heads) are often used to generate multiple sets of dynamic weights for summation, which can preserve additional information by concatenating the results derived from each glimpse. When applying attention mechanisms to an image, image feature vectors that are relevant to different regions are weighted differently to produce an attended image vector. 
\subsubsection{Image attention}
An LSTM model for text question processing was extended by incorporating an image attention model conditioned on previous LSTM hidden states, whose inputs were concatenations of the current word embedding and attended image feature \cite{item43}. The final LSTM hidden state was used as a fused multimodal representation to predict an answer for pointing and grounded VQA. The attention model for an RNN-based encoder-decoder model was used to assign attention weights to image features for image captioning \cite{item44}. Additionally, for VQA, an attention model conditioned on both images and query feature vectors was applied to pinpoint image regions that were relevant to the answer \cite{item45}. 
Similarly, stacked attention networks (SANs) have been proposed to use multiple layers of attention models to query an image multiple times to infer an answer progressively and simulate a multi-step reasoning procedure \cite{item46}. In each layer, a refined query vector is generated and sent to the next layer by adding the previous query vector to the attended image vector produced using the current attention model. A spatial memory network is a multi-hop method for VQA that aligns words to image regions in a first hop and assigns image attention based on entire questions in a second hop to derive an answer \cite{item47}. 

A dynamic memory network is augmented to use separate input modules to encode questions and images. This type of network uses attention-based GRUs to update episodic memory iteratively and retrieve required information \cite{item48}. The bottom-up and top-down attention method (Up-Down), as its name suggests, simulates the human visual system using a combination of two visual attention mechanisms \cite{item49}. The bottom-up attention mechanism proposes a set of salient image regions identified by a Faster R-CNN and the top-down attention mechanism performs concatenation of visual and linguistic features to estimate attention weights and produce an attended image feature vector for image captioning or VQA. The attended image feature vector can be fused with linguistic features again by computing an element-wise product. 
Complementary image features derived from different models, such as ResNet and Faster R-CNN, can be used for multiple image attention mechanisms \cite{item50}. Furthermore, the inverse of image attention, which generates attended text features from image and text inputs, can be used for text-to-image generation \cite{AttnGAN,ObjGAN}.

\subsubsection{Symmetric attention for images and text}
In contrast to the aforementioned image attention mechanisms, co-attention mechanisms use symmetric attention structures to generate not only attended image feature vectors but also attended language vectors \cite{item51}. Parallel co-attention uses a joint representation to derive image and language attention distributions simultaneously. In contrast, alternating co-attention has a cascaded structure that first generates an attended image vector using linguistic features and then generates an attended language vector using the attended image vector. 

Similar to parallel co-attention, a dual attention network (DAN) estimates attention distributions for images and languages simultaneously to derive attended feature vectors \cite{item52}. Such attention models are conditioned on both features and memory vectors related to relevant modalities. This is a key difference compared to co-attention because memory vectors can be iteratively updated at each reasoning step using repeated DAN structures. Memory vectors can be either shared for VQA or modality-specific for image-text matching. 
Stacked latent attention (SLA) improves SANs by concatenating original attended image vectors with values from earlier layers in the attention model to retain latent information from intermediate reasoning stages \cite{item53}. 
A parallel co-attention like twin-stream structure is also included to assign attention to both image and language features, which facilitates iterative reasoning using multiple SLA layers. 
Dual recurrent attention units implement a parallel co-attention structure using LSTM models for text and images to assign attention weights to each input location in representations obtained by convoluting image features using a stack of CNN layers \cite{item54}.
To model high-order interactions between modalities, the high-order correlations between two data modalities can be computed as the inner product of two feature vectors and used to derive attended feature vectors for both modalities \cite{item55}. 

\subsubsection{Attention in a bimodal transformer}
As mentioned in Section~\ref{ssec:representation:transformer}, the bimodal extensions of BERT rely on different tokens to indicate whether a vector is a word or image fragment. Attention models then fuse images with words in bimodal input sequences \cite{item23,item24,item25,item26,item58}. OmniNet uses a gated multi-head attention model in each decoder block to fuse vectors from other modalities with those produced for the current modality by the previous layers in each block \cite{item20}. LXMERT uses independent encoders to learn intra-modality features for each modality and a cross-modality encoder on a higher level to learn cross-modality features using additional cross-attention layers \cite{item21}. ViLBERT extends BERT to include two encoder streams to process visual and textual inputs separately. These features can then interact through parallel co-attention layers \cite{item21}.  

\subsubsection{Other attention-like mechanisms}
The gated multimodal unit is a method that can be viewed as assigning attention weights to images and text based on gating \cite{item56}. This method computes a weighted sum of visual and textual feature vectors based on dimension-specific scalar weights generated dynamically by a gating mechanism. Similarly, element-wise multiplication can be used to fuse visual and textual representations. These fused representations are then used to create the building blocks for a multimodal residual network based on deep residual learning \cite{item57}. 
A dynamic parameter prediction network uses a dynamic weight matrix to transform visual feature vectors, whose parameters are dynamically generated by hashing text feature vectors \cite{Noh2016:DPPnet}. 

\subsection{Bilinear Pooling-based Fusion}
Bilinear pooling is a method that is often used to fuse visual feature vectors with textual feature vectors to create a joint representation space by computing their outer product, which facilitates multiplicative interactions between all elements in both vectors. This method is also referred to as second-order pooling \cite{item59}. In contrast to simple vector combination operations (assuming each vector has $n$ elements), such as a weighted sum, element-wise multiplication, or concatenation, which result in $n$- or $2n$-dimensional representations, bilinear pooling generates an $n^2$-dimensional representation by linearizing the matrix generated by the outer product into a vector, meaning this method is more expressive. Bilinear representations are often linearly transformed into output vectors using a two-dimensional weight matrix, which is equivalent to using a three-dimensional tensor operator to fuse two input feature vectors. Each feature vector can be extended with an extra value of one to preserve single-modal input features in the bilinear representation when calculating an outer product \cite{item60}. However, based on its high dimensionality (typically on the order of hundreds of thousands to a few million dimensions), bilinear pooling often requires the decomposition of weight tensors to allow the associated model to be trained properly and efficiently. 

\subsubsection{Factorization for bilinear pooling}
Because bilinear representations are closely related to polynomial kernels, various low-dimensional approximations can be used to acquire compact bilinear representations \cite{item61}. Count sketches and convolutions can be used to approximate polynomial kernels \cite{item62,item63}, leading to multimodal compact bilinear pooling (MCB) \cite{item64}. Alternatively, by imposing low ranks on weight tensors, multimodal low-rank bilinear pooling (MLB) factorizes three-dimensional weight tensors for bilinear pooling into three two-dimensional weight matrices \cite{item65}. Specifically, visual and textual feature vectors are linearly projected onto low-dimensional modality-specific factors by two input factor matrices. These factors are then fused using element-wise multiplication, followed by linear projection using the third matrix for output factors. 
Multimodal factorized bilinear pooling (MFB) modifies MLB with an extra operation to pool element-wise multiplication results by summing the values within each non-overlapping one-dimensional window \cite{item66}. Multiple MFB models can be cascaded to model high-order interactions between input features, which is referred to as multi-modal factorized high-order pooling (MFH) \cite{item67}. 

MUTAN, which is a multimodal tensor-based Tucker decomposition method, uses Tucker decomposition \cite{item68} to factorize the original three-dimensional weight tensor operator into a low-dimensional core tensor and the three two-dimensional weight matrices used by MLB \cite{item69}. Core tensors model the interactions across modalities. MCB can be considered as MUTAN with fixed diagonal input factor matrices and a sparse fixed core tensor, while MLB can be considered as MUTAN with the core tensor set to the identity tensor. Recently, BLOCK, which is a block-based super-diagonal fusion framework, was proposed to perform block-term decomposition \cite{item70} to compute bilinear pooling \cite{item71}. BLOCK generalizes MUTAN as a summation of multiple MUTAN models to provide richer modeling of the interactions between modalities. MUTAN core tensors can be arranged as super-diagonal tensors, similar to the submatrices of a block diagonal matrix. 
Furthermore, bilinear pooling can be generalized to more than two modalities, such as using outer products to model the interactions among representations for video, audio, and language \cite{item60,item73}. 

\subsubsection{Bilinear pooling and attention mechanisms}
Bilinear pooling can be combined with attention mechanisms. MCB/MLB fused bimodal representations can be used as input features for an attention model to derive an attended image feature vector, which is then fused with a textual feature vector by using MCB/MLB again to form a final joint representation \cite{item64,item65}. MFB/MFH can be used for alternating co-attention to learn joint representations \cite{item66,item67}. A bilinear attention network (BAN) uses MLB to fuse images and text to produce a bilinear attention map representing an attention distribution, which is then used as a weight tensor for bilinear pooling to fuse images and text features again \cite{item74}. 


\section{Applications}
\label{sec:applications}
This section discusses selected applications for multimodal intelligence that combine vision and language, including image captioning, text-to-image generation, and VQA.
Note that there are other common applications, including text-based image retrieval \cite{Lee2018:ECCV} and visual-and-language navigation \cite{Wang2019:CVPR}, that we have not included in this review owing to space limitations.
\begin{itemize}
\item Image captioning is a task that aims to generate a natural language description of an image automatically. It requires a level of image understanding beyond that provided by typical image recognition and object detection methods. 
\item The inverse of image captioning is text-to-image generation, which generates image pixels according to a description or keywords provided by humans. 
\item VQA is related to image captioning. It often takes an image and a free-form, open-ended natural language question about the image as inputs and then outputs a classification result as an answer. Natural language understanding is necessary because questions are free in form. Other capabilities, such as knowledge-based reasoning and commonsense reasoning, are also important because questions are open-ended. 
\item Visual reasoning can be included in all of the aforementioned tasks. However, only methods related to VQA are reviewed here.
\end{itemize}
Detailed task specifications, datasets, and selected work for each task will be introduced in this section.

\subsection{Image Captioning}
Image captioning is a task that requires the generation of a text description of an image ~\cite{he2017deep}. It is one of the first tasks involving the multimodal combination of images and text. 
We mainly review deep learning-based methods. The image captioning task can be divided into several sub-tasks, allowing captions to be generated in a step-by-step manner \cite{fang2015captions, kiros2014unifying, karpathy2015deep}.   
For example, a deep CNN model can be trained to detect words in images, and then a log-linear language model can be used to compose
words into sentences \cite{fang2015captions}. Similarly, image features can be fed into a log-linear language model to generate sentences \cite{kiros2014unifying}. In contrast, exact matching of objects in images and words in sentences attempts to determine if an image and sentence match with each other \cite{karpathy2015deep}. 

Similar to RNN-based encoder-decoder methods for machine translation \cite{Bahdanau2015:NMT}, another approach was 
proposed to generate captions from images in an 
end-to-end manner using an encoder-decoder architecture \cite{vinyals2015show,mao2014deep,chen2015mind}. In this type of model, a CNN, which is typically pre-trained
using ImageNet~\cite{deng2009imagenet}, encodes an image into a continuous vector, which is then fed into an RNN/LSTM decoder to generate captions directly. These types of methods all use the same basic architecture, but they vary slightly in their choices of CNN parameters and how image vectors are fed into decoders. Although this method is powerful and convenient, the encoder-decoder architecture lacks the ability to
capture the fine-grained relationships between objects in images and words in sentences. To overcome this issue, the attention-based
encoder-decoder model has been proposed and has become the standard benchmark for this task~\cite{xu2015show}.
In the attention encoder-decoder model, prior to generating the next word, the decoder first calculates matching scores (attention) with objects in an image and then considers the weighted
image features to generate the next token. There have been many studies that have attempted to improve
the attention model by incorporating additional structures. 
For example, Lu \textit{et al.} added a gate at every decoding step to determine if the next word should
be generated using image information \cite{lu2017knowing}. Additionally, detected words and image features can be combined as inputs 
for the decoder network \cite{gan2017semantic,You2016:cvpr}. More recently, many studies have incorporated extra structures/knowledge from either 
images \cite{item49} or text \cite{deshpande2019fast}. Specifically, an object detector was used to localize features for an image object and generate captions based on
the localized features \cite{item49}. This method improved upon the previous state-of-the-art model by a wide margin in terms of a variety of evaluation metrics.

Image captions with rich information can be generated by incorporating external knowledge. For example, based on a database of celebrities \cite{Guo2016:ECCV}, a the CaptionBot application was developed to describe the components (such as activities) of an image, as well as who is related to each component if the people in the image can be recognized \cite{Tran2016:CVPR}. In addition to generating factual descriptions of images, other approaches have been proposed for explicitly controlling the style \cite{Gan2017:CVPR}, semantic content \cite{gan2017semantic}, and diversity \cite{Li2018:arXiv} of generated captions.

\subsection{Text-to-image Generation}
\label{ssec:application:synthesis}
Text-to-image generation, which relies on natural language to control image generation, is a fundamental problem in computer vision. It is considered to be a difficult problem because it involves at least two tasks: high-quality image generation and language understanding. Generated images must be both visually realistic and semantically consistent with language descriptions. 
Deep learning-based text-to-image generation can be dated back to the use of LSTM for iterative handwriting generation \cite{Graves2013:LSTMSeq}. This iterative image generation method was later extended to create the deep recurrent attentive writer (DRAW) method, which combines an LSTM-based sequential variational auto-encoder (VAE) with a spatial attention mechanism \cite{DRAW}. The alignDRAW method modifies DRAW to use natural language-based descriptions to synthesize images with general content \cite{alignDRAW}. An attention model is used to compute the alignment between input words and iteratively drawn patches. 
GAN-based methods have become the major focus of more recent text-to-image generation studies, potentially because the discriminators of GANs can serve as reasonable criteria for evaluating synthesized images, which is difficult to achieve using other methods. The following subsection provides an overview of some GAN-based methods, including the basic settings and solutions for some important problems, such as the generation of high-quality images, semantic consistency between images and text, and the layout control of images \textit{etc}.

\subsubsection{GAN-based methods}
Compared to VAE, conditional-GAN (CGAN) can synthesize more compelling images of specific categories that a human may even mistake for real images \cite{CGAN,CGANimage}. A GAN model consists of a generator that synthesizes candidates based on input noise and a discriminator that evaluates the candidates. Adversarial training is employed to train the generator to capture true data distributions so the discriminator can no longer discriminate between synthesized data and real data \cite{Goodfellow2014:NIPS}. CGAN extends the standard GAN structure by generating additional category labels for both the generator and discriminator. The GAN-INT-CLS method facilitates the synthesis of visually plausible 64$\times$64 images by using embeddings of natural language descriptions to replace category labels in CGAN \cite{GAN-INT-CLS}. 
Automatic evaluation of the quality of text-conditioned images can be less straightforward. 
To determine the discriminability of GAN-generated images, the inception score \cite{InceptionScore} and Fr\'{e}chet inception distance \cite{FID} metrics are often used. Multi-scale structural similarity \cite{MS-SSIM} is used to evaluate the diversity of images. To evaluate whether a generated image is semantically consistent with an input text description, R-precision \cite{AttnGAN} and visual-semantic similarity \cite{HDGAN} are commonly used metrics. 

\subsubsection{Generating high-quality images}
Although they basically reflect the meanings of descriptions, it has been found that the images produced by GAN-INT-CLS do not contain fine-grained details or vivid objects. This shortcoming motivated the development of the StackGAN method \cite{StackGAN}. StackGAN decomposes image synthesis into more manageable sub-problems through a sketch-refinement process by stacking two separately trained CGANs. The first GAN produces 64$\times$64 low-resolution images by sketching the primitive shapes and colors of objects based on text. The second GAN is then trained to generate 256$\times$256 images by rectifying  defects and adding compelling details to the low-resolution images generated by the first GAN. StackGAN$++$ improves upon this idea by incorporating an additional GAN to generate 128$\times$128 images between the two GANs discussed above and training all GANs jointly
\cite{StackGANv2}. To ensure the generated images semantically match the text precisely, the attentional GAN (AttnGAN) was proposed. This method also stacks three
GANs targeting different image resolutions \cite{AttnGAN}. The first GAN is trained on sentence embeddings, and the next two GANs are trained on bimodal embeddings produced by attention models fusing word-level features with low-resolution images. It has been shown that attention mechanisms can help GANs focus on the words that are the most relevant to the sub-regions drawn in each stage. 
In addition to stacking generators, it has been shown that high-resolution images can be generated using dynamic memory modules \cite{DMGAN}. 
The progressive growth of a GAN begins with training a one-layer generator and one-layer discriminator to synthesize 4$\times$4 images. This method then progressively adds more layers to both models to increase image resolution up to 1024$\times$1024 \cite{ProgGANs}. 

\subsubsection{Generating semantically consistent images}
To improve the semantic consistency between relevant images and text features, DAMSM was proposed for AttnGAN \cite{AttnGAN}. 
The hierarchically-nested discriminator GAN (HDGAN) \cite{HDGAN} handles the same problem by leveraging hierarchical representations with additional adversarial constraints to discriminate not only real/fake image pairs but also real/fake image-text pairs at multiple image resolutions in the discriminator.
Similarly, the text-conditioned auxiliary classifier GAN (TAC-GAN) introduces an additional image classification task into the discriminator \cite{TAC-GAN},
while the text-conditioned semantic classifier GAN (Text-SeGAN) trains classifiers by using regression tasks to estimate the semantic relevance between images and text \cite{Text-SeGAN}. 
As an analogue to cycle consistency \cite{item106}, MirrorGAN was proposed to improve the semantic consistency between two modalities using an additional image captioning module \cite{MirrorGAN}. 

\subsubsection{Semantic layout control for complex scenes}
Despite the success in the generation of realistic and semantically consistent images for single objects, such as birds \cite{Caltech} or flowers \cite{Oxford}, state-of-the-art text-to-image generation methods still struggle to generate complex scenes containing many objects and relationships, such as those in the Microsoft COCO dataset \cite{COCO}. In the pioneering work in \cite{GAWWN}, both text descriptions and the locations of objects specified by keypoints or bounding boxes were used as inputs. Later, detailed semantic layouts, such as scene graphs, were used to replace natural language sentences with more direct descriptions of objects and their relationships \cite{Stanford,UBC}. Additionally, efforts have been made to maintain natural language inputs while incorporating the concept of semantic layouts. Hinz \textit{et al.} included extra object pathways in both a generator and discriminator to control object locations explicitly \cite{Hamburg}. Hong \textit{et al.} employed a two-stage procedure that first constructs a semantic layout automatically from an input sentence using LSTM-based box and shape generators and then synthesizes images using image generators and discriminators \cite{Michigan}. Because fine-grained word/object-level information is  not explicitly used for generation, 
such synthesized images do not contain sufficient details to make them look realistic. The object-driven attentive GAN (Obj-GAN) improves upon the two-stage generation concept using a combination of an object-driven attentive image generator and object-wise discriminator \cite{ObjGAN}.
At every generation step, the generator uses a text description as a semantic layout and synthesizes image regions within a bounding box by focusing on the words that are the most relevant to the object within that box. Obj-GAN is more robust and interpretable compared to other GAN methods and significantly improves object generation quality for complex scenes.

\subsubsection{Additional topics}
In addition to layouts, other types of fine-grained control for image generation have been discussed in the literature. Attribute2Image \cite{Attribute2Image} uses various attributes for face generation, such as age and gender. This concept has also been adapted to face editing to remove beards or change hair colors \cite{AttGAN}. The text-adaptive GAN \cite{TAGAN} facilitates the semantic modification of input images of birds and flowers based on natural language. Lao \textit{et al.} proposed to enforce the learning of representation content and styles as two disentangled variables by using a dual inference mechanism based on cycle-consistency for text-to-image generation \cite{DualGAN}. The success of these methods demonstrates that GANs are able to learn some semantic concepts as disentangled representations, as discussed in Section \ref{ssec:representation:vector}. Text2Scene is another noteworthy method that generates compositional scene representations from natural language in a step-by-step manner without using GANs \cite{Text2Scene}. It has been shown that with minor modifications, Text2Scene can generate cartoon-like, semantic layout, and real image-like scenes. Dialogue-based interactions have also been studied to control image synthesis by improving complex scene generation progressively \cite{ChatPainter,TDR,Chat-crowd,SeqAttnGAN,CoDraw}.
Text-to-image generation has also been extended to multiple images or videos, where visual consistency among generated images is required \cite{StoryGAN,VideoGAN1,VideoGAN2}.

\subsection{Visual Question Answering}
\subsubsection{Task definition}
VQA extends text-based QA from NLP by asking questions related to visual information presented in an image or video clip. Image-based VQA is often considered as a visual Turing test, where a system is required to understand any form of natural language-based questions and answer them in a natural manner. However, it is often simplified as a classification task defined in different ways to focus on different core problems \cite{item43,item84,item85,item86,item87}. Initial works generated questions using templates or by converting descriptive sentences using syntax trees \cite{item84,item89}. Later studies focused on the use of free-form natural language questions authored by either humans or powerful deep generative models, such as GANs and VAEs \cite{item87,item89,item92,item93}. In contrast to open-ended questions, which are presented in complete sentence form, possible answers are often presented as a large set of classes (\textit{e.g.}, 3000 classes) related to yes/no answers, counts, object classes, and instances \textit{etc}. To focus on core understanding and reasoning problems, VQA can be simplified to classify visual and textual features into answer-related classes. 

Alternatively, VQA can be defined to select outputs among multiple (e.g., four) choices, where each choice is associated with an answer presented in the form of a natural language sentence \cite{item43}. This setup can be implemented as a classification problem based on the features of images, questions, and answer candidates \cite{item64}. There are  other types of VQA task definitions as well, such as the Visual Madlibs dataset, which requires answering questions using a ``fill-in-the-blanks'' system \cite{item94}. Furthermore, visual dialogue can be viewed as the answers to a series of questions grounded in images \cite{item95,item96}. This method extends VQA by requiring the generation of more human-like responses and the inference of context based on dialogue history. 

\subsubsection{Common datasets and approaches}
The first VQA data set, which is called DAQUAR, uses real-world images combined with both template-based and human-annotated questions \cite{item84}. COCO-QA contains more QA pairs than DAQUAR because it converts image descriptions from the MS COCO dataset into questions \cite{item89}. Such questions are generally easier to answer because they allow models to rely more on rough images, rather than logical reasoning. VQA v1 and v2 are the most popular datasets for VQA. These datasets consist of open-ended questions with both real and abstract scenes \cite{item87,item97}. A VQA challenge based on these datasets has been held annually as a workshop since 2016. Visual7W is a portion of the Visual Genome dataset for VQA containing multiple choices \cite{item43}. It contains questions related to the concept of ``what'', ``who'', and ``how'' for spatial reasoning and ``where'', ``when'', and ``why'' for high-level commonsense reasoning. The seventh type of questions in Visual7W are ``which'' questions, which are also referred to as pointing questions. The answer choices for these questions are associated with the bounding boxes of objects in images. Approaches designed for these datasets often focus on fusing image and question vectors with the aforementioned discussed attention- and bilinear-pooling-based methods, including SAN, co-attention, Up-Down, MCB, MLB, and BAN \textit{etc}. 

\subsubsection{Integrating external knowledge sources}
Since most of the VQA questions in the aforementioned datasets focus on simple counting, color, and object detection problems that do not require any external knowledge, a possible extension of these tasks is to include more difficult questions that require knowledge beyond what the questions entail or what information is contained in images. Both knowledge-based reasoning for VQA and fact-based VQA datasets incorporate structured knowledge bases, which often require additional steps to query knowledge bases, meaning the corresponding methods are no longer trainable in an end-to-end manner \cite{item98,item99}. In contrast to structured knowledge bases, outside-knowledge VQA uses external knowledge in the form of natural language sentences collected by retrieving Wikipedia articles using search queries extracted from questions. Additionally, an ArticleNet model is trained to find answers in retrieved articles \cite{item100}.

\subsubsection{Discounting language priors}
Although significant achievements have been made, recent studies have pointed out that common VQA benchmarks suffer from strong and prevalent priors (\textit{e.g.}, ``most bananas are yellow'' and ``the sky is mostly blue''), which can often cause VQA models to overfit statistical biases and tendencies in answer distributions. This issue largely circumvents the need to understand visual scenes. Based on the objects, attributes, and relationships provided by the scene graphs of Visual Genome, a new dataset called GQA was created to reduce biases by generating questions using a functional program that controls reasoning steps \cite{item101}. New splits for VQA v1 and VQA v2 were generated to provide different answer distributions for every question in the training and test sets. These splits are referred to as VQA under challenging priors (VQA-CP v1 and VQA-CP v2) \cite{item102}. Other methods have been proposed to handle biased priors using adversarial training or additional training-only structures \cite{item103,item104}. 

\subsubsection{Additional issues}
Another problem that current VQA methods suffer from is low robustness against linguistic variation in questions. 
A dataset called VQA-Rephrasings modifies the VQA v2 validation set with human-authored rephrasing of questions \cite{item106}. 
Additionally, a cycle-consistency-based \cite{item107} method was proposed to improve linguistic robustness by enforcing consistency between original and rephrased questions, as well as between true answers and answers predicted based on original and rephrased questions. 
Zhang \textit{et al.} suggested that attention mechanisms can cause VQA models to suffer from counting object proposals and an additional model component was proposed as a solution \cite{item108}. Additionally, it is known that current VQA methods cannot read text from images. A method was proposed to address this problem by fusing text extracted from images using optical character recognition  \cite{item109}. VizWiz is a goal oriented VQA dataset collected by blind people capturing potentially low-quality images and asking questions in spoken English. This dataset includes many text-related questions \cite{item110}. 
To learn rare concepts that humans may talk more likely than the commonsense knowledge, active learning, which allows a model to seek labels selectively for more informative examples, has been applied to VQA to reduce data annotation efforts \cite{Lin2017:VQA,Misra2018:LBA,Jedoui2019:VQA}.

\subsection{Visual Reasoning}
This section focuses on the study of a very intriguing problem called visual reasoning, which focuses on how to accomplish accurate, explicit, and expressive understanding and reasoning. Visual reasoning is related to many language- and vision-based bimodal tasks, such as captioning and text-to-image generation. However, in this section, we mostly focus on methods related to VQA because visual reasoning is particularly important when answering complicated questions.
SANs are often considered as being closely related to implicit visual reasoning because their stacked structures can be viewed as performing multiple reasoning steps. Feature-wise linear modulation was proposed to refine visual features iteratively using feature-wise affine transformations based on scaling factors and bias values generated dynamically from textual features \cite{item83}.
Multimodal relational networks (MuRel) also have structures with multiple MuRel cells based on bilinear pooling, which can
be used iteratively \cite{item72}.

\subsubsection{Neural module network-based methods}
A neural module network (NMN) consists of a collection of jointly trained neural ``modules'' combined into a deep model for answering questions \cite{item111}. A dependency parser first helps convert natural language questions into a fixed and rule-based network format and specifies both the set of modules used to answer questions and the connections between modules. Next, a deep model is assembled based on the target question format to generate answer predictions. SHAPES, which is a synthetic dataset consisting of complex questions regarding simple arrangements of ordered shapes, was proposed to focus on the compositional aspects of questions \cite{item111}. A later study trained a model layout predictor jointly with module parameters by re-ranking a list of layout candidates using reinforcement learning. This method is called a dynamic NMN \cite{item112}. Modules for ``find'' or ``relate'' operations use attention models to focus on one or two regions in an input image and make the execution of assembled deep models similar to running a functional program \cite{item112}. An end-to-end version of the NMN used an RNN question encoder to convert input questions into layout policies without requiring the aid of a parser \cite{item113}. This work was based on a relatively new dataset called compositional language and elementary visual reasoning diagnostics (CLEVR). As its name suggests, CLEVR is a synthetic diagnostic dataset for testing a range of visual reasoning abilities related to objects and relationships with minimal biases and detailed annotations describing the type of reasoning each question requires \cite{item114}. 
Other implementations of the NMN include the program generator and execution engine method (PG+EE), which shares generic designs among certain operations \cite{item115}; the stack-NMN, which improves the parser and incorporates question features into modules \cite{item116}; and the transparency-by-design network, which redesigns some modules from PG+EE to maintain the transparency of the reasoning procedure \cite{item117}.

\subsubsection{Other types of end-to-end reasoning methods}
Another end-to-end approach is the memory, attention, and composition (MAC) network, which decomposes questions into a series of attended reasoning steps and performs each step using a recurrent MAC cell that maintains a separation between control and memory hidden states. Each hidden state is generated by an ANN model constructed based on attention and gating mechanisms \cite{item118}. 
Recently, both deterministic symbolic programs and probabilistic symbolic models have been used as execution engines for generated programs to improve transparency and data efficiency, resulting in the creation of the neural-symbolic VQA (NS-VQA) and probabilistic neural-symbolic models, respectively \cite{item119,item120}. As an extension of the NS-VQA, the neuro-symbolic concept learner (NS-CL) uses a neuro-symbolic reasoning module to execute programs based on scene representations. The NS-CL can have its program generator, reasoning module, and visual perception components trained jointly in an end-to-end manner without requiring any component-level supervision \cite{item121}. Its perception module learns visual concepts based on language descriptions of objects and facilitates learning new words and parsing new sentences. 

We conclude this section by reviewing the relationship network (RN), which has a simple structure that uses an ANN as a function to model the relationships between any pair of visual and textual features. The resulting output values are then accumulated and transformed by another ANN
\cite{RelationNetworks}. Although the RN simply models relationships without any form of inductive reasoning, it achieves very high VQA accuracy on the CLEVR dataset. This inspires a re-thinking of the connections between correlation and induction.

\section{Summary and Prospects}
This paper reviewed the topics of modeling and machine learning across multiple modalities based on deep learning with a focus on the combination of vision and natural language.
We organized many different works from the language-vision multimodal intelligence field according to three factors: multimodal representations, fusion of multimodal signals, and applications of multimodal intelligence. In the section on representations, both single-modal and multimodal representations were reviewed based on the key concept of embedding. Multimodal representations unify relevant signals from different modalities into the same vector space for general downstream tasks. For multimodal fusion, special architectures, such as attention mechanisms and bilinear pooling, were discussed. In the application section, three selected areas of broad interest were presented: image captioning, text-to-image generation, and VQA. A set of visual reasoning methods for VQA was also discussed. Our review covered task definition, dataset specification, development of commonly used methods, as well as issues and trends. We hope this review will promote future studies in the emerging field of multimodal intelligence.


In the future, in addition to the aforementioned research topics, we also want to highlight the following three directions.

\subsection{Multimodal Knowledge Learning}
Multiple knowledge bases related to multimodal datasets have been constructed in recent years, such as MS-Celeb-1M \cite{MSCeleb1M}, which benchmarks recognition of one million celebrities in images and links them to their corresponding information in freebase \cite{freebase}.  
In this area, the automatic acquisition of commonsense knowledge from multimodal data can be expected in the near future. 
Massive amounts of information, including entities, actions, attributes, concepts, and relationships, can be learned from massive amounts of image and video data to construct models covering broad and structured commonsense knowledge. Such models will provide great value for applications related to commonsense reasoning. 
However, problems that need to be resolved to accomplish this goal include the following:
\begin{itemize}
\item Defining commonsense;
\item Constructing multimodal datasets and learning commonsense knowledge from them efficiently and effectively;
\item Determining which tasks to work on to verify the capabilities of novel algorithms while demonstrating the importance of commonsense; 
\item Updating previously learned commonsense knowledge.
\end{itemize}  

\subsection{Multimodal Emotional Intelligence}
Advanced emotional intelligence is a cognitive ability unique to humans. Communication between humans involves rich emotions and multiple modalities. To construct a highly anthropomorphic human–-computer interaction agent, machines must understand and generate multimodal emotional content and empathize with humans. 
Fundamental research 
in this area can not only help us to understand the mechanisms of cognitive intelligence, but also has great value for many real-world applications. 
However, the difficulties of multimodal emotional intelligence include the following:
\begin{itemize}
\item Perceiving and aligning the subtle expression of emotions in different modalities;
\item Ensuring the consistency and rationality of data across all modalities \cite{HanJing}; 
\item Acquiring the core representations and intensities of emotions that are potentially modality-invariant \cite{Chaturvedi2019:PRL} .
\end{itemize}

\subsection{Large-scale Complex Goal-oriented Multimodal Intelligent Human-computer Interaction System}
The intelligentization of service industries is both a large opportunity and large technical challenge for artificial intelligence. Considering e-commerce as an example, this field faces challenges related to ultra-large-scale data and complex human-computer interactions in the full retail chain. These problems require large-scale, complex, and task-oriented multimodal intelligent human-computer interaction technologies to serve hundreds of millions of users in a personalized and highly efficient way. 
To this end, opportunities exist in terms of promoting open-source and open-license frameworks for multimodal human-–computer interaction systems, constructing large-scale datasets and algorithm verification platforms, and conducting fundamental research on multimodal intelligence. Breakthroughs related to techniques in these areas will also promote the intelligentization of broader service industries.

Regarding the goal of constructing an agent that can perceive multimodal information and use the connections between different modalities to improve its cognitive ability, research on multimodal intelligence is still in its infancy. However, it has already achieved significant progress and become a very important branch of the development of artificial intelligence.

\section*{Acknowledgement}
The authors are grateful to the editor and anonymous reviewers for their valuable suggestions that helped to make this paper better.
This work is partially supported by Beijing Academy of Artificial Intelligence (BAAI). 

\ifCLASSOPTIONcaptionsoff
  \newpage
\fi



%

{\scriptsize
\bibliography{mybib}{}
\bibliographystyle{ieeetr}
}

\end{document}